\documentclass{amii}

\usepackage{amsmath,amsfonts,bm}

\def\eqref#1{equation~\ref{#1}}

\def\1{\bm{1}}

\DeclareMathAlphabet{\mathsfit}{\encodingdefault}{\sfdefault}{m}{sl}
\SetMathAlphabet{\mathsfit}{bold}{\encodingdefault}{\sfdefault}{bx}{n}

\usepackage{booktabs}
\usepackage{hyperref}
\usepackage{url}
\usepackage{subcaption}
\usepackage{graphicx}
\usepackage{adjustbox}

\hypersetup{hidelinks}
\usepackage{tikz}
\usepackage[scaled=0.95]{helvet}

\usetikzlibrary{arrows.meta, positioning, calc}

\usepackage{algorithm}
\usepackage{algpseudocode}

\title{Sparse RL Relaxation for LLM Reasoning}

\author[1,2]{Jiabin Fan}
\author[4]{Matthias Lindemann}
\author[5]{Zita Marinho}
\author[4,5,6]{André FT Martins}
\author[1,2,3]{Lili Mou}

\affiliation[1]{University of Alberta}
\affiliation[2]{Alberta Machine Intelligence Institute (Amii)}
\affiliation[3]{Canada CIFAR AI Chair}
\affiliation[4]{Instituto de Telecomunicações}
\affiliation[5]{Instituto Superior Técnico, Universidade de Lisboa}
\affiliation[6]{ELLIS Unit Lisbon}

\abstract{Recent work has explored continuous representations in place of discrete tokens during Chain-of-Thought (CoT) reasoning, often referred to as soft CoT. This is motivated by the idea that a mixture of token embeddings can represent a superposition of multiple reasoning possibilities within a single step. However, it remains challenging to train large language models to realize this potential.
Existing work has explored reinforcement learning (RL)-style training for soft CoT. However, these approaches treat soft tokens as actions in a Markov decision process rather than as differentiable intermediate computation steps. Thus, they do not propagate gradients across the soft tokens and miss out on the signal of how the choice of one soft token affects the next. In this paper, we propose a sparse relaxation method for RL (called SpaRLax), which leverages the differentiability of soft-CoT reasoning over time steps and makes training more stable and effective. We adopt an entmax sparsity method that adaptively determines the degree of softness, which truncates long-tail noise at confident steps while retaining multiple plausible candidates when uncertain. Across nine reasoning benchmarks, our method achieves the highest average performance compared with baseline methods. Further analyses validate the effective use of
soft-token superposition in our approach, and also demonstrate the contribution of adaptive sparsity in our approach.}

\date{\today}

\begin{document}

\maketitle

\section{Introduction}

Chain of Thoughts \citep[CoT;][]{wei2022chain} improves the reasoning ability of large
language models (LLMs) by generating intermediate steps before the final answer.
Despite its effectiveness, CoT produces a single discrete token at each step, collapsing the LLM's output distribution to one token and discarding alternatives that may be useful later. This has motivated recent work on \textit{soft CoT}, where intermediate
reasoning tokens are represented by continuous embeddings \citep{hao2025coconut,wang2025softconcept,zhang2025softthinking,wu2025pccot,shen2025codi,xu2025softcot}. Among these approaches, \citet{zhang2025softthinking} construct each CoT token as a probability-weighted mixture of token embeddings, referred to as a \emph{soft token}, allowing multiple candidate semantics to coexist at a reasoning step.
Theoretical analyses  \citep{zhu2025superposition} further show that such continuous representation can represent multiple reasoning paths in superposition and can be more expressive than discrete CoT, suggesting soft CoT as a promising direction for improving LLM reasoning.

However, training LLMs to realize the potential of soft reasoning remains challenging.
Recent methods struggle to fit soft CoT training into the reinforcement learning (RL) framework, which allows us to learn soft CoT without groundtruth labels. To facilitate RL, researchers have introduced various stochasticity mechanisms to soft CoT, such as Gaussian embedding perturbations~\citep{butt2026softtokens} and Gumbel-softmax sampling~\citep{zheng2025softgrpo}.
However, such a formulation is problematic, as the output-generating process defines the vocabulary as the action space\footnote{In the standard token-level MDP for autoregressive LLMs, the policy is defined on
vocabulary tokens $a_i\in V$ as $\pi_{\bm\theta}(a_i| s_i)$ for some state $s_i$.
A soft token $\tilde {\bm e}_i\notin V$ is not in this action space, so a
policy-gradient objective over soft tokens is ill-defined under standard RL frameworks for LLMs~\citep{ramamurthy2023rl}.}
but soft 
reasoning considers the embedding as an action. They do not coexist in the same Markov decision process (MDP), making the action space, and thus the entire RL formulation, ill-defined. In addition, such a formulation fails to take advantage of the differentiable nature of soft tokens, as the gradient is not back-propagated across the soft tokens. This misses out on the signal of how the choice of one soft token affects the next.

In this paper, we propose a Sparse RL Relaxation (SpaRLax) method to train soft reasoning for LLMs. For every reasoning step, we sample a mixture probability by Gumbel-softmax~\citep{jang2017gumbel, maddison2017concrete}, which introduces stochasticity compared with directly using softmax. Different from \cite{zheng2025softgrpo}, however, we notice that Gumbel-softmax is differentiable, and with this insight, we restore the cross-step gradient path, allowing gradients to flow backward from each soft token to the one before it. 
Moreover, we notice an additional challenge that the typical large vocabulary size of an LLM dilutes probability mass, causing a loss of reasoning focus. We propose a solution to this problem by using sparse relaxation, which concentrates probability mass on a few candidate tokens, while assigning exactly zero probability to low-scoring tokens. In particular, we apply entmax
\citep{peters2019sparse} on top of the Gumbel reparametrization trick
when constructing the soft-token distribution, as
entmax can naturally adapt the degree of softness by the uncertainty of each
reasoning step.

We conduct experiments on nine benchmarks spanning mathematical,
knowledge-intensive, and code reasoning. We compare our method against soft-CoT post-training methods under a controlled GRPO framework. The results show that our SpaRLax achieves the strongest weighted-average
performance under both soft- and hard-CoT inference, demonstrating the benefits
of cross-step learning in the relaxed reasoning space
(\S\ref{sec:main_results}). 
We also observe that our approach 
utilizes the superposition encoded by soft tokens more
than existing methods (\S\ref{sec:superposition_analysis}). In addition, we
analyze the sparsity mechanism across different candidate-set sizes and find
that entmax outperforms dense softmax, suggesting its effectiveness in
adaptively suppressing long-tail noise (\S\ref{sec:sparsity_analysis}).

\section{Preliminaries}
\label{sec:background}
\paragraph{Classic CoT reasoning.}
In classic CoT reasoning \citep{wei2022chain}, we generate a sequence of discrete reasoning tokens, typically enclosed by special tokens such as
$\langle\mathrm{think}\rangle$ and $\langle/\mathrm{think}\rangle$, before
producing the final answer.
We denote an input prompt by~$\bm x$, the reasoning tokens by $\bm t_{1:m}=(t_1,\ldots,t_m)$, and the final answer tokens by $\bm y_{1:n}=(y_1,\ldots,y_n)$. The LLM's autoregressive sequence is given by
\begin{align}
    \bm x
    &\rightarrow
    \langle\mathrm{think}\rangle
    \:
    \underbrace{t_1,\cdots,t_m}_{\text{discrete CoT}}
    \:
    \langle/\mathrm{think}\rangle
    \:
    \underbrace{y_1,\cdots,y_n}_{\text{final answer}}.
    \label{eq:cot_trajectory}
\end{align}
At the $i$th reasoning step, 
the LLM with its policy $\pi_{\bm \theta}(\cdot|\cdot)$ yields a probability vector
$\bm p_i \in \Delta^{|\mathcal V|-1}$ over the vocabulary $\mathcal V$
and samples a token $t_i$ by
\begin{align}
    \bm  p_i
    &=
    \pi_{\bm \theta}
    \left(
        \cdot | \bm x,\bm t_{<i}
    \right),
    \qquad
    t_i \sim \mathrm{Categorical}(\bm p_i).
    \label{eq:discrete_inference}
\end{align}
Once $t_i$ is sampled, its embedding, denoted by $\operatorname{emb}(t_i)$, is passed to the next step, while the remaining information in $\bm p_i$ is not carried forward.
After $\langle/\mathrm{think}\rangle$, the LM switches to
answer generation.

\paragraph{Soft CoT reasoning.}
\citet{zhang2025softthinking} propose soft CoT, which replaces the
discrete token $t_i$ in classic CoT with a continuous soft-token embedding
to improve CoT expressivity.
In this paradigm, the model predicts the soft-token distribution
$\tilde{\bm  p}_i$ and constructs the soft-token embedding by
\begin{align}
    \tilde {\bm  p}_i 
    &=
\pi_{\bm{\theta}}\left(\cdot | \bm{x},\tilde{ \bm{e}}_{<i}\right),
    \qquad
    \tilde {\bm e}_i
    =
    \sum_{k=1}^{|\mathcal V|}
    \tilde p_{i,k}\operatorname{emb}(k),
    \label{eq:soft_inference}
\end{align}
where $\tilde p_{i,k}$ is the $k$th element of $\tilde{\bm p}$.
The resulting $\tilde {\bm e}_i$ is fed back to the model for the next step until $\langle/\mathrm{think}\rangle$ is generated (details can be seen in Appendix~\ref{app:stopping_criterion}). Then the LLM generates discrete tokens for the final answer, and the entire autoregressive sequence becomes
\begin{align}
    \bm x
    &\rightarrow
    \langle\mathrm{think}\rangle
    \:
    \underbrace{
        \tilde {\bm e}_1,\tilde {\bm e}_2,\cdots,\tilde {\bm e}_m
    }_{\text{soft CoT}}
    \:
    \langle/\mathrm{think}\rangle
    \:
    \underbrace{
        y_1,y_2,\cdots,y_n
    }_{\text{(discrete) final answer}}.
    \label{eq:soft_cot_trajectory}
\end{align}

\paragraph{RL training for LLMs.}
LLMs are often post-trained with reinforcement learning (RL) to improve reasoning ability, where the model explores different reasoning trajectories and is rewarded for trajectories that lead to correct answers.
For a prompt $\bm{x}$, the model samples a trajectory including CoT
$t_{1:m}$ and final answer $y_{1:n}$. Then, the model receives a scalar reward
$R(\bm x,\bm y_{1:n})$ for an entire trajectory, based on the correctness of the final answer for the input, but typically not based on CoT steps.
RL optimizes the policy parameter $\bm{\theta}$ to maximize the expected reward:
\begin{align}
    J(\bm{\theta})
    &=
    \mathbb{E}_{\bm x\sim\mathcal D,\,
    \bm t_{1:m},\bm y_{1:n}\sim\pi_{\bm\theta}}
    \left[
        R(\bm x,\bm y_{1:n})
    \right],
    \label{eq:rl_objective}
\end{align}
where $\mathcal D$ denotes the distribution of training prompts.

Because discrete token sampling is non-differentiable, policy-gradient
methods~\citep{williams1992reinforce,sutton2000policy} can be used to differentiate
Eqn.~(\ref{eq:rl_objective}):
\begin{equation}
\label{eq:discrete_pg_gradient}
\begin{adjustbox}{max width=\linewidth}
$\displaystyle
\nabla_{\bm\theta} J(\bm\theta)
=
\mathop{\mathbb E}\limits_{
    \substack{
        \bm x\sim\mathcal D,\\
        \bm t_{1:m},\,\bm y_{1:n}\sim\pi_{\bm\theta}
    }
}
\left[
    R(\bm x,\bm y_{1:n})
    \left(
        \sum_{i=1}^{m}
        \nabla_{\bm\theta}
        \log
        \pi_{\bm\theta}
        \left(
            t_i
            |
            \bm x,\bm t_{<i}
        \right)
        +
        \sum_{j=1}^{n}
        \nabla_{\bm\theta}
        \log
        \pi_{\bm\theta}
        \left(
            y_j
            |
            \bm x,\bm t_{1:m},\bm y_{<j}
        \right)
    \right)
\right].
$
\end{adjustbox}
\end{equation}
Recent studies extend the likelihood-based formulation
to soft reasoning.
Let $\rho_{\bm\theta}(\tilde{\bm e}_i |\bm x,\tilde {\bm e}_{<i})$ denote the likelihood of a sampled soft token by the Gaussian- \citep{butt2026softtokens} or Gumbel-based \citep{zheng2025softgrpo} stochastic 
mechanism. Their policy gradient takes the form
\begin{equation}
\label{eq:soft_pg_gradient}
\begin{adjustbox}{max width=\linewidth}
$\displaystyle
\nabla_{\bm\theta} J(\bm\theta)
=
\mathop{\mathbb E}\limits_{
    \substack{
        \bm x\sim\mathcal D,\\
        \tilde{\bm e}_{1:m}\sim\rho_{\bm\theta},\,
        \bm y_{1:n}\sim\pi_{\bm\theta}
    }
}
\left[
    R(\bm x,\bm y_{1:n})
    \left(
        \sum_{i=1}^{m}
        \nabla_{\bm\theta}
        \log
        \rho_{\bm\theta}
        \left(
            \tilde{\bm e}_i
            |
            \bm x,\tilde{\bm e}_{<i}
        \right)
        +
        \sum_{j=1}^{n}
        \nabla_{\bm\theta}
        \log
        \pi_{\bm\theta}
        \left(
            y_j
            |
            \bm x,\tilde{\bm e}_{1:m},\bm y_{<j}
        \right)
    \right)
\right].
$
\end{adjustbox}
\end{equation}

Such methods view the soft token defined by
Eqn.~(\ref{eq:soft_inference}) as a continuous RL action, which is less justified under
the standard Markov decision process (MDP) framework for autoregressive LLMs. This is because an LLM treats discrete vocabulary tokens as the action space, but it becomes unclear how a soft token embedding is compatible with discrete actions of tokens in the vocabulary. In addition, soft tokens could have been modeled as differentiable variables, but treating them as RL actions breaks this differentiable structure.
Hence, these methods miss out on the gradient signal of how the choice of one soft token affects the later reasoning steps.

\section{Our Method: Sparse RL Relaxation (SpaRLax)
}
\label{sec:method}
In this section, we introduce the proposed SpaRLax approach in detail. In \S\ref{sec:rl_relaxation}, we discuss our RL relaxation based on Gumbel-softmax; we also present an efficient approximation for implementation considerations. Then, in \S\ref{sec:sparse_rl_relaxation} we introduce sparsity by replacing  Gumbel-softmax with Gumbel-\emph{entmax}, yielding sparse soft tokens whose degree of softness adapts across reasoning steps.

\subsection{RL Relaxation}
\label{sec:rl_relaxation}

Soft CoT naturally forms a differentiable autoregressive computation across reasoning steps. We exploit this structure and propose an RL relaxation method. At reasoning step $i$, the LLM $f_{\bm \theta}$ produces pre-softmax logits 
\begin{align}
    \bm z_i
    =
    f_{\bm\theta}
    \left(
        \bm x,\tilde{\bm e}_{<i}
    \right)
    \in
    \mathbb{R}^{|\mathcal V|}.
    \label{eq:relax_logits}
\end{align}
To introduce stochastic exploration, we perturb each logit with i.i.d.~Gumbel
noise $\epsilon_{i,k}\sim\mathrm{Gumbel}(0,1)$.
By the Gumbel-Max trick \citep{maddison2014astar}, the argmax of the perturbed
logits is an exact categorical sample from
$\operatorname{softmax}(\bm z_i)$. For differentiability, the argmax operation is replaced with softmax, resulting in the Gumbel-softmax operation~\citep{jang2017gumbel,maddison2017concrete}. For the next step $i+1$, the input is a soft token embedding, given by
\begin{align}
    \tilde{\bm p_i}
    &=
    \operatorname{softmax}
    \left(
        \frac{
            \bm z_i+\bm\epsilon_i
        }{\tau_g}
    \right),
    \qquad
    \tilde{\bm e}_{i+1}
    =
    \sum_{k=1}^{|\mathcal V|}
    \tilde p_{i,k}\operatorname{emb}(k),
    \label{eq:rl_relaxation}
\end{align}
where $\tau_g>0$ is the relaxation temperature \citep{goyal2017differentiable}. 
As seen, given sampled $\bm\epsilon_i$, the relaxed soft token $\tilde{\bm e}_i$ remains differentiable with
respect to $\bm z_i$. 
\begin{figure*}[t!]
    \centering
    \vspace{0.2cm}\includegraphics[width=0.96\textwidth]{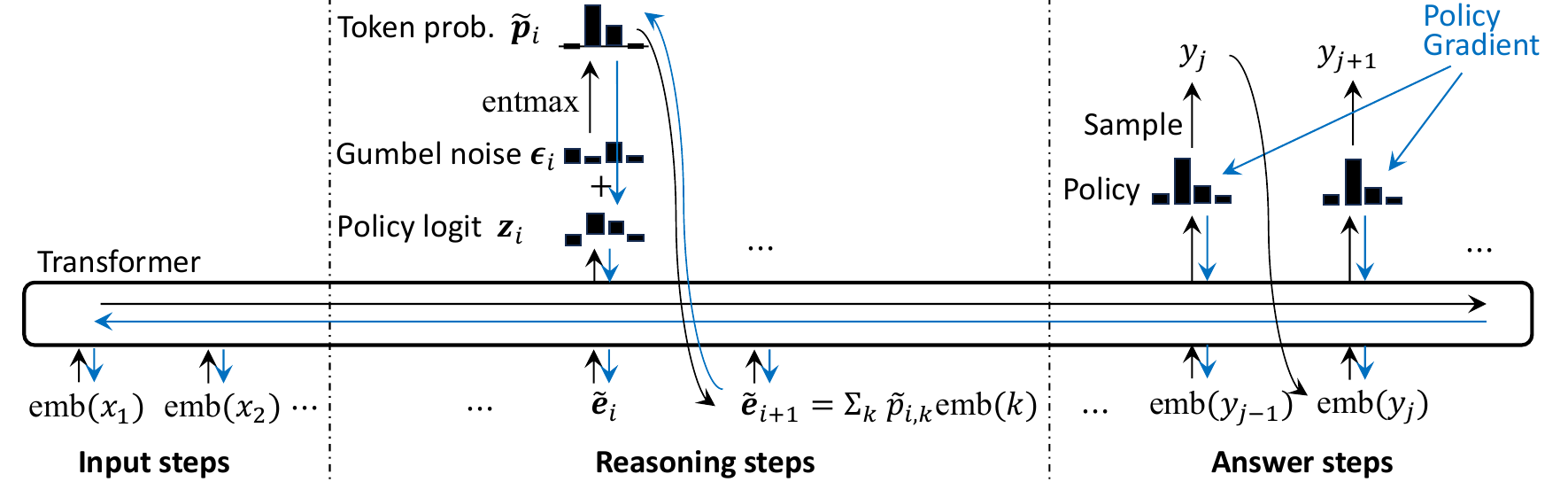}
    \vspace{0.3cm}
    \caption{
     Overview of our approach. \textbf{Black} arrows indicate forward propagation, and {\color{blue} \textbf{blue}} arrows indicate backpropagation. Different from Soft-GRPO~\citep{zheng2025softgrpo}, we only apply policy gradient to output steps, but backpropagate cross-step gradients for reasoning steps.
    }
    \label{fig:gradient_flow}
\end{figure*}

After the reasoning steps, the LLM generates a discrete output  $\bm y_{1:n}$ by sampling from its policy. The training objective is to maximize the expected reward:
\begin{align}
    J_{\mathrm{relax}}(\bm\theta)
    =
    \mathbb{E}_{
        \substack{
            \bm x\sim\mathcal D,\,
            \bm \epsilon_{1:m}\sim \operatorname{Gumbel}(0,1)\\
            \bm y_{1:n}\sim
            \pi_{\bm\theta}
            (\cdot
            |
            \bm x,\tilde{\bm e}_{1:m})
        }
    }
    \left[
        R(\bm x,\bm y_{1:n})
    \right]
    \label{eq:relaxed_expected_reward}
\end{align}
In other words, Gumbel-softmax allows us to sample from unparametrized Gumbel noise, eliminating the parameters underneath the expectation operation. This facilitates backpropagation through the reasoning steps. The
gradient of Eqn.~(\ref{eq:relaxed_expected_reward}) can be written as
\begin{equation}
\label{eq:relax_gradient}
\begin{adjustbox}{max width=\linewidth}
$\displaystyle
\nabla_{\bm\theta}
J_{\text{relax}}(\bm\theta)
=
\mathop{\mathbb E}\limits_{
    \substack{
        \bm x\sim\mathcal D,\,
        \bm \epsilon_{1:m}\sim \operatorname{Gumbel}(0,1)\\
        \bm y_{1:n}\sim
        \pi_{\bm\theta}
        \left(
            \cdot
            |
            \bm x,\tilde{\bm e}_{1:m}
        \right)
    }
}
\left[
    R(\bm x,\bm y_{1:n})
    \sum_{j=1}^{n}
    \nabla_{\bm\theta}
    \log
    \pi_{\bm\theta}
    \left(
        y_j
        |
        \bm x,
        \tilde{\bm e}_{1:m},
        \bm y_{<j}
    \right)
\right],
$
\end{adjustbox}
\end{equation}
where the policy gradient applies to the discrete answer tokens $\bm y_{1:m}$ only, which differs from previous work \citep{zheng2025softgrpo,butt2026softtokens} treating soft reasoning tokens as actions. As illustrated in Fig.~\ref{fig:gradient_flow}, the reasoning trajectory
remains differentiable through the relaxed soft-token construction, allowing
the learning signal to propagate across reasoning steps.

\textbf{Efficient approximation for implementation considerations.} 
Modern RL implementations \citep{sheng2025hybridflow, fu2025areal} adopt a two-pass strategy: (1) The first pass performs roll-out by generating an entire trajectory without backpropagation. This does not require us to store activation values of the neural network and is memory-efficient, but it is slow because tokens are generated autoregressively.
(2) The second pass takes in the sampled actions for gradient calculation, which is memory-intensive but fast. The two-pass strategy is essential for performing RL on LLMs, as it allows us to perform the roll-out and policy update (of different batches) simultaneously, so as to maximize GPU parallelization \citep{fu2025areal}. 

However, this presents a challenge for our backpropagation. Note that the sampled soft token embeddings are generated in the first pass, but are fed as input in the second pass, thus detached from the computational graph. Consequently, there is no gradient flow from $\tilde {\bm e}_{i}$ to $\tilde{\bm p}_{i-1}$, $\tilde{\bm z}_{i-1}$, and $\tilde{\bm e}_{i-1}$.

Inspired by surrogate-gradient methods
\citep{bengio2013estimating,wang2021scalable}, we propose a \textit{proxy gradient grafting} technique that connects the gradient from $\tilde {\bm e}_{i}$ to its previous step. The key idea is to preserve the cached rollout values in the forward pass,
while recomputing each soft-token distribution under the current policy and
using its gradient in the backward pass.

Let
$(\tilde{\bm p}_i^{\text{roll}},
  \tilde{\bm e}_i^{\text{roll}}, \bm \epsilon_i)$
denote the soft-token distribution and embedding cached from rollout,
together with the sampled Gumbel noise $\bm\epsilon_i$. In a traditional implementation, the logits $\bm z_i$ are computed from $\tilde{\bm e}_i^\text{roll}$. However, we reconstruct it from  $\tilde{\bm e}_{i-1}^\text{roll}$ at one-step earlier:
\begin{align}
    \bm z_{i-1}^{\text{proxy}}
    &=
    f_\theta
    \left(
        \bm x,
            \tilde{\bm e}_{\le i-1}^{\text{roll}}
    \right),
    \qquad
    \tilde{\bm p}_{i-1}^{\text{proxy}}
    =
    \operatorname{softmax}
    \left(
        \frac{
            \bm z_{i-1}^{\text{proxy}}+\bm\epsilon_{i-1}
        }{\tau_g}
    \right).
    \label{eq:proxy_distribution}
\end{align}
Here, we reuse the same $\bm\epsilon_{i-1}$ sampled during rollout, so we have $\bm z_{i-1}^\text{proxy}=\bm z_{i-1}^\text{roll}$. Then, we graft the proxy gradient onto the cached one for calculating the soft-token embedding for the next step.
\begin{align}
    \bar{\bm p}_{i-1}
    &=
    \operatorname{sg}
    \left(
        \tilde{\bm p}_{i-1}^{\text{roll}}
    \right)
    +
    \tilde{\bm p}_{i-1}^{\text{proxy}}
    -
    \operatorname{sg}
    \left(
        \tilde{\bm p}_{i-1}^{\text{proxy}}
    \right),
    \qquad
    \bar{\bm e}_{i}
    =
    \sum_{k=1}^{|\mathcal V|}
    \bar p_{i-1,k}\operatorname{emb}(k),
    \label{eq:proxy_grafting}
\end{align}
where $\operatorname{sg}(\cdot)$ denotes stop gradient of a variable. Then, $\bar{\bm e}_{i}$ replaces $\tilde{\bm e}_{i}^\text{roll}$ in the standard second pass of the policy update. Note that the cached variable $\tilde{\bm p}_{i-1}^\text{roll}$ is already detached due to the two-pass implementation, but we explicitly add the $\operatorname{sg}$ operation in the formula for clarity.

With this construction, the forward computation exactly preserves the cached roll-out trajectory, as we always have $\bar {\bm p}_{i-1}=  \tilde{\bm p}_{i-1}^{\text{roll}}$. However, we can now let the gradient flow from $\bar{\bm e}_i$ to $\tilde{\bm p}_{i-1}^{\text{proxy}}$ by standard backpropagation. 

In terms of efficiency, our grafting technique does not have high latency, as we only graft one earlier step. Therefore, Eqns.~(\ref{eq:proxy_distribution}) and (\ref{eq:proxy_grafting}) can be computed in parallel for all time steps $i$. However, this grafting trick has a limitation, as the gradient backpropagates to only one earlier step from $\bar{\bm e}_i$ to $\tilde{\bm p}_{i-1}^{\text{proxy}}$ (for each value of $i$). That said, all steps can be adequately trained because the attention layers backpropagate gradients to all earlier steps. Moreover, we argue that our cross-step gradient is important even for one step. This is because the nearest step receives the strongest signal, as the gradient tends to vanish as the number of steps increases in recurrent computation~\citep{bengio1994longterm,pascanu2013difficulty}.

Our method provides meaningful signal on how one soft token should affect the next.

\textbf{Compatibility with RL post-training objectives.}
Our RL relaxation is independent of the particular policy-optimization rule used
for RL post-training.
It modifies how the soft-CoT computation is constructed and differentiated,
while the RL policy optimization is an outer layer often involving advantage estimation,
importance weighting, and policy-update mechanism.
Thus, our RL relaxation can be combined with commonly used post-training
algorithms such as PPO \citep{schulman2017ppo}, GRPO
\citep{shao2024deepseekmath}, DAPO \citep{yu2025dapo}, GSPO
\citep{zheng2025gspo}, and REINFORCE++ \citep{hu2025reinforcepp}.

\subsection{Sparse RL Relaxation}
\label{sec:sparse_rl_relaxation}

Prior soft-CoT methods
\citep{zhang2025softthinking,zheng2025softgrpo,zhou2026lepo}
use softmax to map LLM logits into a soft-token distribution, assigning
non-zero probability to every vocabulary token. This mixes a long tail of low-probability
tokens into soft token embeddings, which introduces unnecessary noise aggregated over
a large vocabulary. A common remedy is top-$k$ truncation, which limits the mixture to a fixed number of highest-scoring tokens. However, a fixed $k$ cannot accommodate the varying uncertainty across reasoning steps. As observed by \citet{fu2026selar}, token distributions can be highly concentrated at some steps but much more diffuse at others. Thus, a large fixed $k$ may retain unnecessary noise, while a small $k$ may discard useful alternatives.

To suppress this long-tail noise while retaining multiple candidates
when necessary, we replace the softmax relaxation in
Eqn.~(\ref{eq:rl_relaxation}) with $\alpha$-$\operatorname{entmax}$
\citep{peters2019sparse}, yielding an adaptive sparse RL Relaxation.
For logits $\bm z\in\mathbb{R}^{|\mathcal V|}$, entmax is defined by
\begin{align}
    \operatorname{entmax}_{\alpha}(\bm z)_k
    =
    \left[
        (\alpha-1)z_k-\lambda
    \right]_{+}^{\frac{1}{\alpha-1}},
    \label{eq:entmax}
\end{align}
where $\lambda$ normalizes the distribution and $\alpha>1$ controls its sparsity, with
$\alpha \to 1$ recovering $\operatorname{softmax}$ and $\alpha=2$ corresponding to $\operatorname{sparsemax}$ \citep{martins2016softmax}. 
For $\alpha>1$, entmax assigns exactly zero probability to sufficiently weak
tokens, while the number of retained tokens is determined by their relative scores.
Accordingly, Sparse RL Relaxation constructs the soft-token distribution as
\begin{align}
    \tilde{\bm p}_i
    &=
    \operatorname{entmax}_{\alpha}
    \left(
        \frac{
            \bm z_i+\bm\epsilon_i
        }{\tau_g}
    \right),
    \qquad
    \tilde{\bm e}_{i+1}
    =
    \sum_{k=1}^{|\mathcal V|}
    \tilde p_{i,k}\operatorname{emb}(k).
    \label{eq:gumbel_entmax}
\end{align}
If only one logit dominates (after Gumbel noise in our scenario), $\operatorname{entmax}$ suppresses the remaining tokens, and the soft token approaches a discrete representation.
When several candidates have comparable logits, it retains them as a sparse
mixture.
Thus, the degree of softness adapts to the uncertainty of each reasoning
step without requiring a manually specified support size.

The same replacement can be applied when reconstructing
$\tilde{\bm p}_i^{\text{proxy}}$ during proxy gradient grafting.
Since $\operatorname{entmax}$  remains differentiable on its active support, gradients can propagate
through the sparse soft-token construction in the same manner as
in \S\ref{sec:rl_relaxation}.

\section{Experiments}
\label{sec:exp}

\subsection{Setup}
\label{sec:exp_setup}

\begin{table*}[t]
\centering
\setlength{\tabcolsep}{3pt}
\renewcommand{\arraystretch}{1.13}

\resizebox{\textwidth}{!}{%
\begin{tabular}{lcccccccccc}
\toprule
& \multicolumn{6}{c}{Math Reasoning}
& \multicolumn{1}{c}{Knowledge}
& \multicolumn{2}{c}{Code}
& \\
\cmidrule(lr){2-7}
\cmidrule(lr){8-8}
\cmidrule(lr){9-10}

Method
& GSM8K
& MATH
& AIME24
& AIME25
& AMC23
& Olym.
& GPQA
& HEval
& MBPP
& \textbf{W.Avg.} \\
\midrule

\multicolumn{11}{l}{\textit{Hard CoT Inference}} \\[1pt]

Initial
& 80.08 {\scriptsize$\pm$ 0.73}
& 75.18 {\scriptsize$\pm$ 0.96}
& 15.94 {\scriptsize$\pm$ 4.12}
& 20.94 {\scriptsize$\pm$ 2.90}
& 59.69 {\scriptsize$\pm$ 4.95}
& 26.24 {\scriptsize$\pm$ 0.75}
& 20.55 {\scriptsize$\pm$ 2.61}
& 55.11 {\scriptsize$\pm$ 2.19}
& 58.55 {\scriptsize$\pm$ 1.87}
& 59.58 {\scriptsize$\pm$ 0.48} \\

Discrete-GRPO
& \underline{84.42} {\scriptsize$\pm$ 0.72}
& 75.97 {\scriptsize$\pm$ 0.95}
& \underline{16.98} {\scriptsize$\pm$ 4.26}
& 21.25 {\scriptsize$\pm$ 2.84}
& \underline{62.11} {\scriptsize$\pm$ 5.82}
& 27.31 {\scriptsize$\pm$ 0.74}
& 21.26 {\scriptsize$\pm$ 2.23}
& 56.23 {\scriptsize$\pm$ 2.31}
& \textbf{59.45} {\scriptsize$\pm$ 2.14}
& \underline{61.90} {\scriptsize$\pm$ 0.38} \\

Soft-GRPO
& 84.29 {\scriptsize$\pm$ 0.55}
& 75.48 {\scriptsize$\pm$ 1.38}
& 15.94 {\scriptsize$\pm$ 4.46}
& 20.94 {\scriptsize$\pm$ 3.51}
& 58.91 {\scriptsize$\pm$ 5.35}
& 26.25 {\scriptsize$\pm$ 0.87}
& 20.15 {\scriptsize$\pm$ 1.90}
& 55.22 {\scriptsize$\pm$ 2.84}
& 57.66 {\scriptsize$\pm$ 1.88}
& 61.24 {\scriptsize$\pm$ 0.47} \\

STHT
& 80.35 {\scriptsize$\pm$ 0.58}
& 76.49 {\scriptsize$\pm$ 1.02}
& 13.85 {\scriptsize$\pm$ 4.65}
& \underline{22.92} {\scriptsize$\pm$ 3.28}
& \underline{62.11} {\scriptsize$\pm$ 5.65}
& \underline{27.36} {\scriptsize$\pm$ 0.75}
& \underline{22.77} {\scriptsize$\pm$ 2.30}
& 55.16 {\scriptsize$\pm$ 2.00}
& 58.18 {\scriptsize$\pm$ 2.04}
& 60.28 {\scriptsize$\pm$ 0.43} \\

LEPO
& 83.36 {\scriptsize$\pm$ 1.26}
& \underline{76.98} {\scriptsize$\pm$ 1.83}
& \textbf{19.58} {\scriptsize$\pm$ 6.07}
& 21.64 {\scriptsize$\pm$ 4.34}
& 61.22 {\scriptsize$\pm$ 7.45}
& 26.73 {\scriptsize$\pm$ 1.34}
& 22.38 {\scriptsize$\pm$ 3.23}
& \underline{56.62} {\scriptsize$\pm$ 4.07}
& \underline{59.41} {\scriptsize$\pm$ 3.45}
& 61.21 {\scriptsize$\pm$ 1.23} \\

SpaRLax (ours)
& \textbf{85.16} {\scriptsize$\pm$ 0.53}
& \textbf{78.24} {\scriptsize$\pm$ 1.09}
& 16.77 {\scriptsize$\pm$ 3.92}
& \textbf{22.97} {\scriptsize$\pm$ 3.16}
& \textbf{68.83} {\scriptsize$\pm$ 4.31}
& \textbf{29.19} {\scriptsize$\pm$ 0.76}
& \textbf{27.43} {\scriptsize$\pm$ 2.59}
& \textbf{57.28} {\scriptsize$\pm$ 2.32}
& 58.82 {\scriptsize$\pm$ 1.89}
& \textbf{63.43} {\scriptsize$\pm$ 0.42} \\

\midrule
\multicolumn{11}{l}{\textit{Soft CoT Inference}} \\[1pt]

Initial
& 80.59 {\scriptsize$\pm$ 0.65}
& 78.92 {\scriptsize$\pm$ 1.09}
& 16.98 {\scriptsize$\pm$ 4.96}
& 24.11 {\scriptsize$\pm$ 3.99}
& 67.97 {\scriptsize$\pm$ 5.06}
& 29.75 {\scriptsize$\pm$ 0.80}
& 32.62 {\scriptsize$\pm$ 3.03}
& 58.38 {\scriptsize$\pm$ 2.03}
& 60.70 {\scriptsize$\pm$ 2.40}
& 62.33 {\scriptsize$\pm$ 0.49} \\

Discrete-GRPO
& 84.93 {\scriptsize$\pm$ 0.69}
& \underline{79.80} {\scriptsize$\pm$ 1.00}
& 16.35 {\scriptsize$\pm$ 4.67}
& \textbf{25.05} {\scriptsize$\pm$ 3.24}
& 68.59 {\scriptsize$\pm$ 3.96}
& 30.61 {\scriptsize$\pm$ 0.69}
& 32.67 {\scriptsize$\pm$ 2.76}
& \textbf{59.98} {\scriptsize$\pm$ 3.10}
& \textbf{61.52} {\scriptsize$\pm$ 1.77}
& \underline{64.58} {\scriptsize$\pm$ 0.46} \\

Soft-GRPO
& \underline{84.96} {\scriptsize$\pm$ 0.57}
& 79.15 {\scriptsize$\pm$ 1.46}
& \underline{17.81} {\scriptsize$\pm$ 5.26}
& 23.18 {\scriptsize$\pm$ 4.07}
& 66.56 {\scriptsize$\pm$ 4.39}
& 30.34 {\scriptsize$\pm$ 0.66}
& 33.43 {\scriptsize$\pm$ 2.67}
& 59.78 {\scriptsize$\pm$ 2.21}
& 60.60 {\scriptsize$\pm$ 1.64}
& 64.35 {\scriptsize$\pm$ 0.45} \\

STHT
& 80.45 {\scriptsize$\pm$ 0.73}
& 79.43 {\scriptsize$\pm$ 1.32}
& 17.19 {\scriptsize$\pm$ 4.33}
& \underline{24.32} {\scriptsize$\pm$ 3.85}
& \underline{68.98} {\scriptsize$\pm$ 5.11}
& \underline{30.63} {\scriptsize$\pm$ 0.90}
& \textbf{34.77} {\scriptsize$\pm$ 2.31}
& 59.17 {\scriptsize$\pm$ 1.83}
& 60.35 {\scriptsize$\pm$ 2.07}
& 62.70 {\scriptsize$\pm$ 0.41} \\

LEPO
& 83.25 {\scriptsize$\pm$ 1.65}
& 76.33 {\scriptsize$\pm$ 1.57}
& 11.88 {\scriptsize$\pm$ 8.99}
& 17.19 {\scriptsize$\pm$ 6.48}
& 61.64 {\scriptsize$\pm$ 7.48}
& 28.99 {\scriptsize$\pm$ 1.72}
& 28.36 {\scriptsize$\pm$ 5.89}
& 56.04 {\scriptsize$\pm$ 4.56}
& 59.03 {\scriptsize$\pm$ 3.25}
& 62.00 {\scriptsize$\pm$ 0.71} \\

SpaRLax (ours)
& \textbf{86.34} {\scriptsize$\pm$ 0.72}
& \textbf{80.01} {\scriptsize$\pm$ 1.11}
& \textbf{18.33} {\scriptsize$\pm$ 4.23}
& 23.39 {\scriptsize$\pm$ 4.55}
& \textbf{69.84} {\scriptsize$\pm$ 4.71}
& \textbf{31.20} {\scriptsize$\pm$ 0.74}
& \underline{34.72} {\scriptsize$\pm$ 2.98}
& \underline{59.82} {\scriptsize$\pm$ 2.53}
& \underline{60.83} {\scriptsize$\pm$ 1.89}
& \textbf{65.42} {\scriptsize$\pm$ 0.51} \\

\bottomrule
\end{tabular}%
}

\caption{
Pass@1 performance (\%) under hard and soft chain-of-thought (CoT) inference.
Results are reported as mean $\pm$ standard deviation.
\textbf{Bold} and \underline{underlined} values indicate the best and
second-best mean performance within each inference setting, respectively.
}
\label{tab:main_pass1_app}
\end{table*}
\begin{table*}[t]
\centering

\scriptsize
\setlength{\tabcolsep}{3.6pt}
\renewcommand{\arraystretch}{1.10}

\begin{tabular}{lccccccccc c}
\toprule
& \multicolumn{6}{c}{Math Reasoning}
& \multicolumn{1}{c}{Knowledge}
& \multicolumn{2}{c}{Code}
& \\
\cmidrule(lr){2-7}
\cmidrule(lr){8-8}
\cmidrule(lr){9-10}

Method
& GSM8K
& MATH
& AIME24
& AIME25
& AMC23
& Olym.
& GPQA
& HEval
& MBPP
& \textbf{W.Avg.} \\
\midrule

\multicolumn{11}{l}{\textit{Hard CoT Inference}} \\[1pt]

Initial
& \textbf{97.50}
& 91.60
& \textbf{50.00}
& 55.00
& 87.50
& 42.07
& 82.32
& 83.54
& \underline{86.77}
& 81.22 \\

Discrete-GRPO \citep{guo2025deepseekr1}
& \textbf{97.50}
& \underline{91.80}
& \underline{46.67}
& \textbf{58.33}
& \underline{95.00}
& 42.96
& 82.32
& \textbf{84.76}
& \textbf{87.16}
& 81.65 \\

SofT-GRPO \citep{zheng2025softgrpo}
& \underline{97.35}
& \underline{91.80}
& 40.00
& 53.33
& 90.00
& 41.19
& 74.75
& 81.71
& 84.05
& 80.14 \\

STHT \citep{butt2026softtokens}
& 97.12
& 91.60
& \underline{46.67}
& \underline{56.67}
& \textbf{97.50}
& 42.37
& \underline{85.35}
& 83.54
& 84.82
& 81.28 \\

LEPO \citep{zhou2026lepo}
& 97.19
& \textbf{92.20}
& \underline{46.67}
& 53.33
& \underline{95.00}
& \underline{43.41}
& \textbf{90.40}
& 82.32
& 86.38
& \underline{81.90}\\

SpaRLax (ours)
& \underline{97.35}
& \underline{91.80}
& \textbf{50.00}
& \textbf{58.33}
& \underline{95.00}
& \textbf{44.15}
& \textbf{90.40}
& \underline{84.15}
& 85.60
& \textbf{82.21} \\

\midrule
\multicolumn{11}{l}{\textit{Soft CoT Inference}} \\[1pt]

Initial
& \underline{97.35}
& \textbf{93.60}
& 46.67
& \underline{60.00}
& \underline{95.00}
& \underline{46.52}
& \underline{96.97}
& 83.54
& 85.99
& 83.38 \\

Discrete-GRPO \citep{guo2025deepseekr1}
& 97.12
& 93.00
& \underline{50.00}
& \textbf{61.67}
& \underline{95.00}
& \underline{46.52}
& 94.95
& \underline{85.37}
& 87.16
& 83.32 \\

SofT-GRPO \citep{zheng2025softgrpo}
& 97.27
& \underline{93.40}
& \textbf{56.67}
& 58.33
& 92.50
& 45.33
& 96.46
& \textbf{85.98}
& \underline{87.55}
& 83.32 \\

STHT \citep{butt2026softtokens}
& \textbf{97.42}
& \textbf{93.60}
& 46.67
& \textbf{61.67}
& \textbf{97.50}
& 45.93
& \textbf{97.47}
& 84.76
& 85.60
& \underline{83.41} \\

LEPO \citep{zhou2026lepo}
& 97.27
& 91.40
& 30.00
& 48.33
& 92.50
& 43.41
& 92.42
& 82.32
& \underline{87.55}
& 81.75 \\

SpaRLax (ours)
& \underline{97.35}
& 93.00
& \underline{50.00}
& 58.33
& \underline{95.00}
& \textbf{47.11}
& 96.46
& 84.15
& \textbf{87.94}
& \textbf{83.56} \\
\bottomrule
\end{tabular}
\caption{
Pass@32 performance (\%) under hard and soft chain-of-thought (CoT) inference.
\textbf{Bold} and \underline{underlined} values denote the best and
second-best results within each inference setting, respectively.
}

\label{tab:main_pass32}
\end{table*}
\textbf{Models and datasets.}
We use DeepSeek-R1-Distill-Qwen-1.5B\footnote{\url{https://huggingface.co/deepseek-ai/DeepSeek-R1-Distill-Qwen-1.5B}}
as the base LLM for all experiments.
We perform RL training on
DAPO-MATH-17k
\citep{yu2025dapo}, a collection of mathematical reasoning problems, which is selected to reduce potential overlap with the
base model's pretraining corpus, allowing the effect of RL fine-tuning to be
evaluated cleanly.

We evaluate the trained models on nine benchmarks spanning multiple reasoning
domains.
For mathematical reasoning, we use GSM8K \citep{cobbe2021training},
MATH \citep{hendrycks2021measuring}, AIME 2024, AIME 2025, AMC 2023,
and OlympiadBench \citep{he2024olympiadbench}.
We further evaluate general knowledge and scientific reasoning on
GPQA-Diamond \citep{rein2024gpqa}, and code reasoning on
HumanEval \citep{chen2021evaluating} and MBPP \citep{austin2021programming}. We compute the weighted average
performance based on the number of data points in each test dataset.

\textbf{Competing approaches.}
We compare our SpaRLax method against representative discrete- and
soft-reasoning baselines.
\textbf{Initial} denotes the base model without training.
\textbf{Discrete-GRPO} applies standard GRPO
\citep{shao2024deepseekmath} to discrete-token CoT.
For RL with soft reasoning, we compare against
\textbf{SofT-GRPO} \citep{zheng2025softgrpo},
\textbf{STHT} \citep{butt2026softtokens},
and \textbf{LEPO} \citep{zhou2026lepo}.
To ensure a controlled comparison, we implement all methods within the
same GRPO framework, meaning that
SofT-GRPO, STHT, LEPO, and our method modify only the components required by their respective soft-CoT learning mechanisms.
This experimental protocol minimizes performance differences caused by the choice of the underlying RL algorithm and allows the experiments to focus on the
effectiveness of the soft-CoT stage. Appendix~\ref{app:sparse_grpo} provides the pseudo-code of our SpaRLax under the GRPO framework.

\textbf{Training and evaluation settings.} 
During training, our SpaRLax uses soft tokens for reasoning steps, but for the answer steps, hard tokens are generated by top-$p$ sampling (with $p=0.95$) in conjunction with top-$k$ tokens ($k=5$), which is standard in previous work~\citep{zheng2025softgrpo,zhou2026lepo}. 
After training, our model supports either soft CoT or hard CoT (simply sampling discrete tokens for training steps). More configurations can be found in Appendix~\ref{app:imp_configs}. 

For evaluation, we report Pass@1 (single-sample accuracy)
and Pass@32 (accuracy of whether at least one correct solution is found among 32 sampled responses). They are commonly used in previous literature~\cite{,,}. To account for the variance of Pass@1 induced by stochastic sampling, we repeat each evaluation four times from the same checkpoint and report the mean and standard deviation. 

\subsection{Main Results}
\label{sec:main_results}

\begin{figure*}[t!]
\centering
\begin{subfigure}[t]{0.35\textwidth}
\centering
\includegraphics[width=\linewidth]{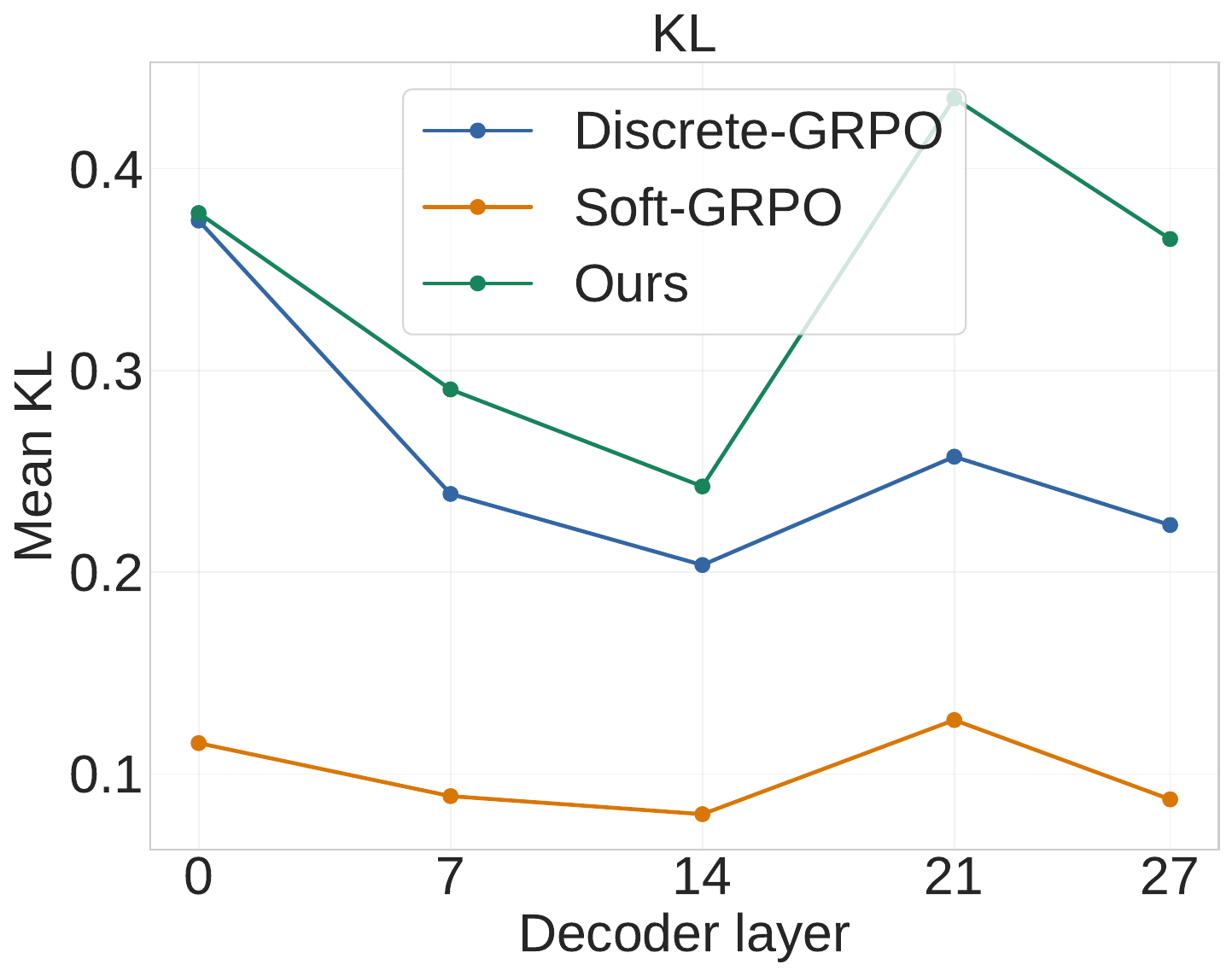}
\end{subfigure}
\quad\quad\quad\quad
\begin{subfigure}[t]{0.35\textwidth}
\centering
\includegraphics[width=\linewidth]{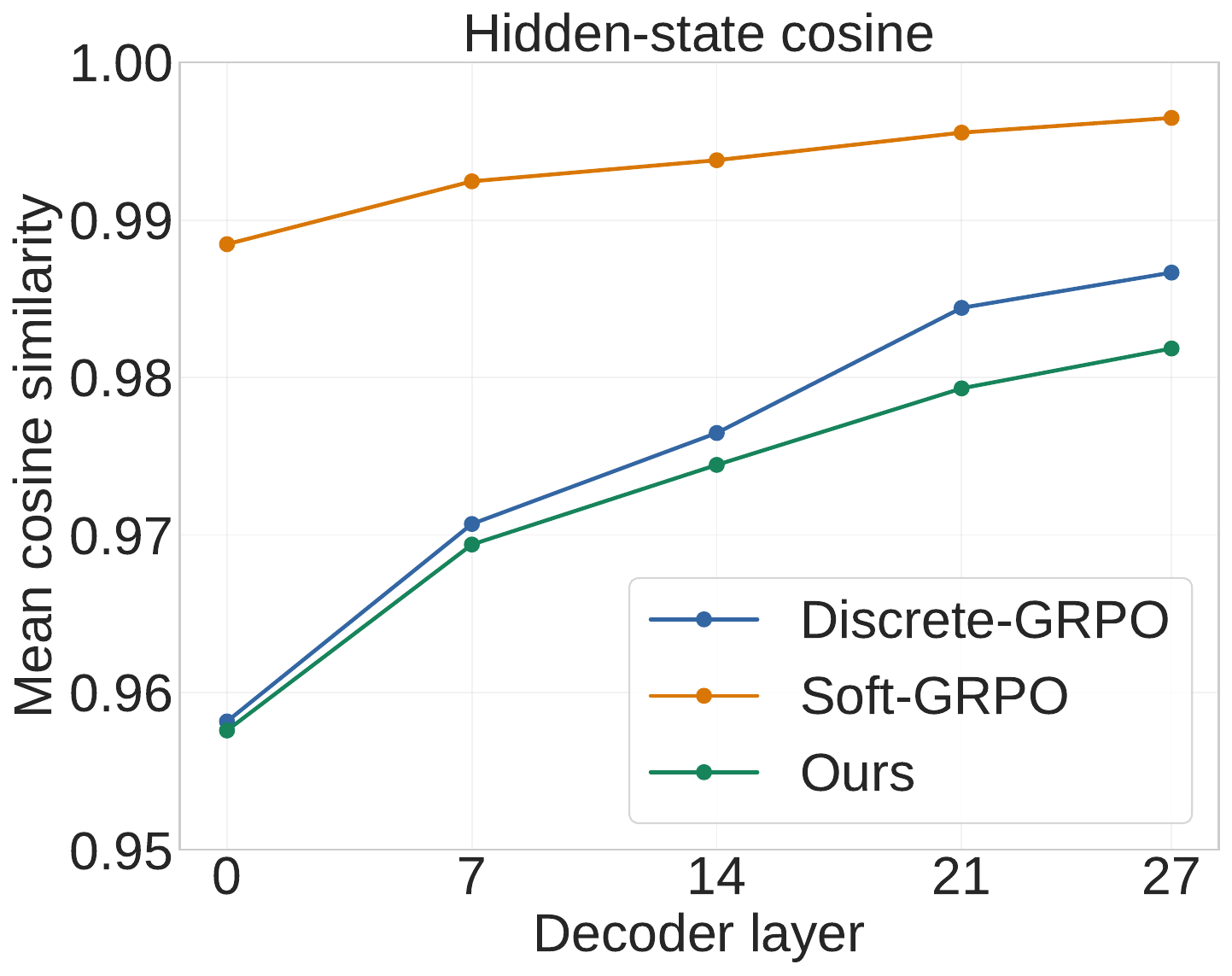}
\end{subfigure}
\caption{Comparing the hidden states when using the soft token embedding and replacing it with top-1 token embedding. Higher KL and lower cosine indicate a stronger utilization of superposition.}
\label{fig:layerwise-superposition}
\vspace*{-5pt}
\end{figure*}
Based on the above setups, we conduct post-training
from the same initial model, select the checkpoint with the best validation
performance, and evaluate each model with both hard-CoT and soft-CoT inference. We report Pass@1 and Pass@32 in Tabs.~\ref{tab:main_pass1_app}--\ref{tab:main_pass32}.

\textbf{Pass@1.} We observe that soft-CoT inference on the initial model already provides a
notable improvement over hard-CoT inference, consistent with prior
work~\citep{zhang2025softthinking}. This advantage persists after
post-training, as soft-CoT inference
generally achieves higher Pass@1 than hard-CoT inference across different post-training methods. Meanwhile, 
soft CoT post-training also brings notable gains under hard-CoT
inference, consistent with prior observations~\citep{butt2026softtokens}.

Our method achieves the strongest Pass@1 under both inference modes, reaching
$63.43$ with hard CoT and $65.42$ with soft CoT. Soft
CoT provides a richer training space and more
expressive inference representation~\citep{zhu2025superposition}. Our method propagates the gradient of reward across soft tokens, providing a learning signal on how the choice of one soft token affects the next. This enables more effective training for soft
reasoning in such a richer training space. As a result, SpaRLax achieves strong improvements under both
soft CoT and hard CoT inference.

\textbf{Pass@32.}
Pass@32 is much higher than Pass@1 for all methods, which is
expected as Pass@32 only
requires one successful solution among 32 stochastic rollouts.
Soft CoT inference continues to outperform hard CoT
inference at Pass@32, consistent with observations in prior
work~\citep{zheng2025softgrpo,butt2026softtokens}. Although the gain of our method is much smaller than those observed for Pass@1, other 
post-training methods yield even smaller Pass@32 improvements over the initial
model. This behavior is in line with prior observations that, under moderate RL
budgets, post-training improves sampling efficiency but does not
expand large-$k$ solution coverage~\citep{yue2025does,liu2025prorl}. That being said, our method achieves the largest average improvement over the initial model among all competing methods.

\subsection{Superposition Utilization}
\label{sec:superposition_analysis}

\citet{rizvimartel2026illusion} show that recent soft reasoning methods often collapse their superposed representation toward a single
token, indicating the observed gain may be associated simply with more greedy reasoning.
In this subsection, we analyze if our method suffers from the same collapse or actually makes use of the
multiple alternatives encoded in soft tokens.

Following \citet{rizvimartel2026illusion}, we examine whether the
information beyond the top-1 component of the soft token persists in the model's hidden states.
For each data sample and each reasoning step $i$, we keep all
preceding soft tokens $\tilde{\bm e}_{<i}$ unchanged and perform two forward
passes that differ only at the current position: one uses the original soft
token $\tilde{\bm e}_i$, while the other replaces it with its top-1 token's embedding.
This isolates the effect of the non-top-1 components at a single reasoning
step. At each Transformer layer, we compare the resulting hidden states using cosine
similarity and symmetric KL divergence between their Logit-Lens
distributions.\footnote{We project an intermediate
hidden state $\bm h_i^{(\ell)}$ to the vocabulary space as
$\bm p_i^{(\ell)}
=\operatorname{softmax}(\bm U\bm h_i^{(\ell)})$,
where $\bm U$ is the output unembedding matrix and $\ell$ denotes the index of transformer layer. Then, we compute the symmetric
KL divergence between the distributions obtained from the soft-token and
top-1-token forward passes.}
We compute these metrics for every inference step of every sample,
and then average them across inference steps and data samples for each layer.
The resulting layerwise averages are plotted in
Fig.~\ref{fig:layerwise-superposition}.
Higher KL divergence and lower cosine similarity indicate that the non-top-1
components have a stronger effect on subsequent computation, indicating a better utilization of the soft tokens in superposition.

As shown in Fig.~\ref{fig:layerwise-superposition}, our method exhibits
larger KL divergence and lower cosine similarity than SofT-GRPO and discrete-GRPO across decoder
layers, suggesting stronger utilization of the information retained in the
soft-token mixture.

\subsection{Effect of Adaptive Sparsity}
\label{sec:sparsity_analysis}

\begin{figure*}[t!]
    \centering
    \begin{subfigure}[t]{0.49\textwidth}
        \centering
        \includegraphics[width=\linewidth]{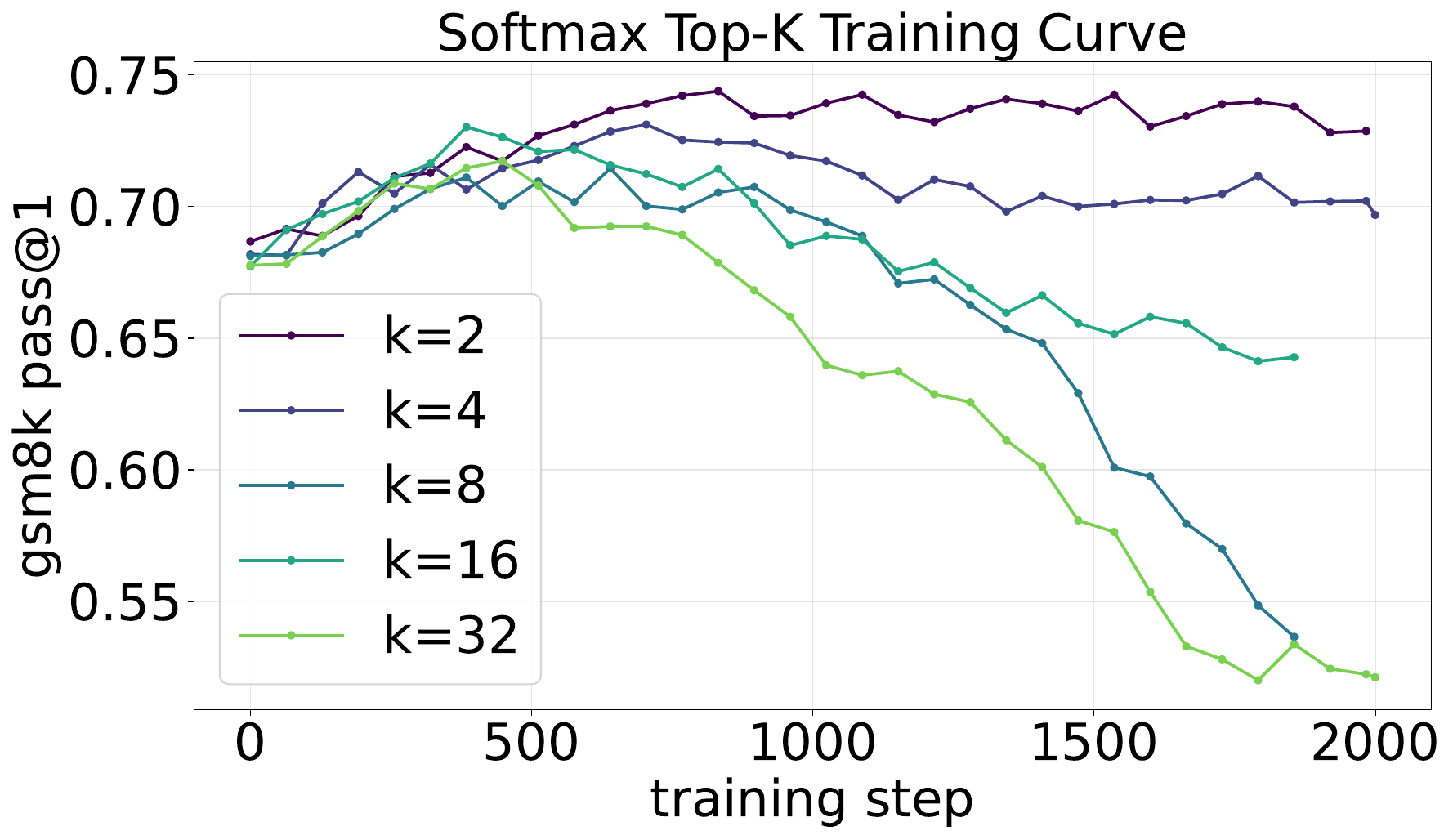}
    \end{subfigure}
    \hfill
    \begin{subfigure}[t]{0.49\textwidth}
        \centering
        \includegraphics[width=\linewidth]{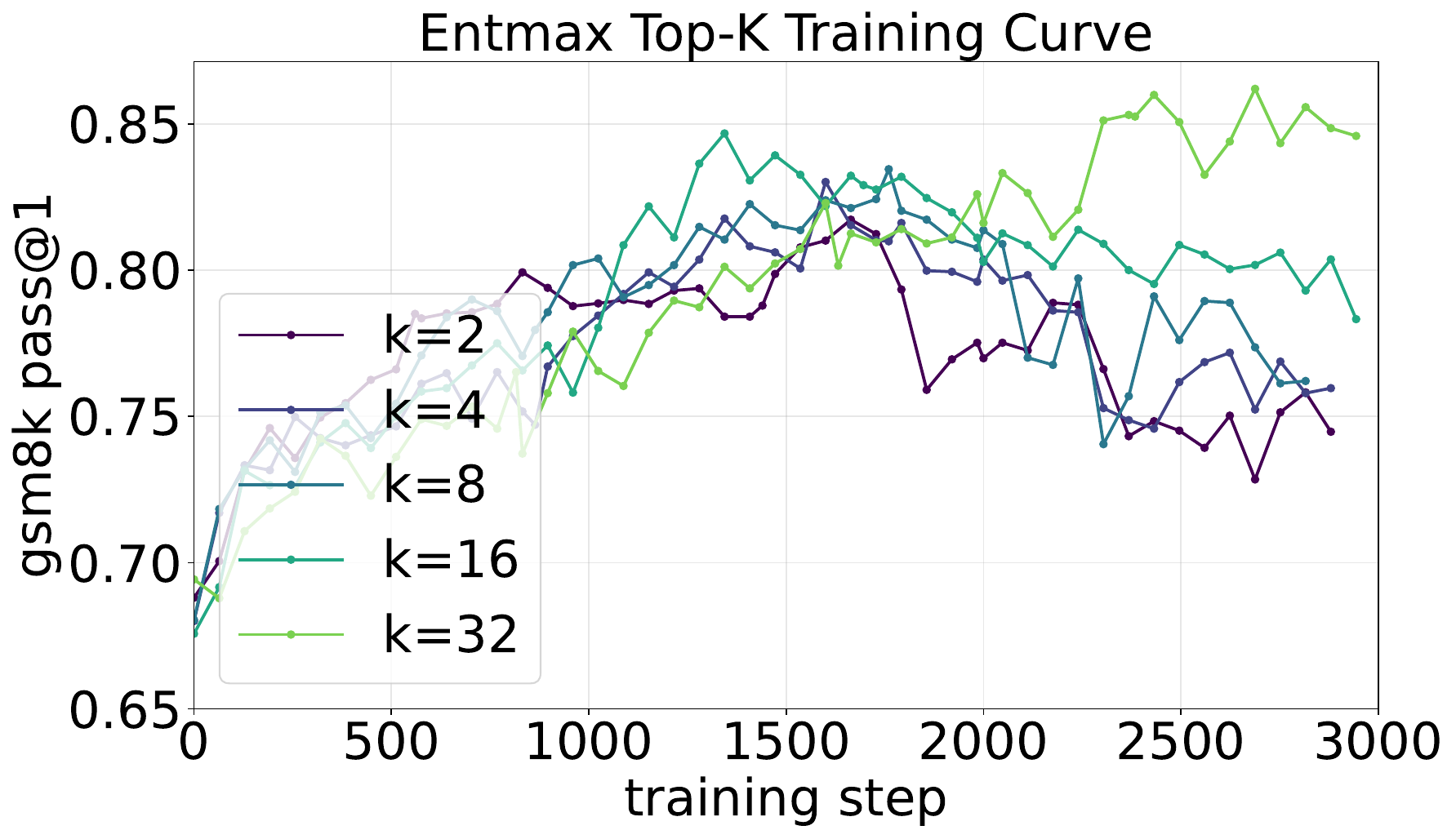}
    \end{subfigure}

    \caption{Effect of candidate-set size on softmax and entmax training.
Softmax becomes increasingly sensitive to larger $k$, whereas entmax
remains robust as $k$ increases.
    }
    \label{fig:topk-ablation}
\end{figure*}
Constructing a soft token over a large vocabulary requires aggregating
all token embeddings, which introduces a huge memory burden and long-tail noise.
Recent methods apply top-$k$
truncation, retaining only the highest-scoring $k$ tokens before constructing
the soft-token embedding
\citep{zheng2025softgrpo,zhou2026lepo}.
While a small $k$ reduces both computational cost and long-tail noise, it also
imposes a limit on the number of alternatives retained at every
reasoning step.
This can restrict the superposition capability of soft tokens when more
candidates are useful.

Our intuition is that the degree of softness should be step-dependent.
As observed in prior work \citep{fu2026selar}, token distributions are 
concentrated at most reasoning steps but highly uncertain at a
small fraction of steps.
Thus, confident steps may require little superposition, whereas
uncertain steps can benefit from retaining multiple plausible alternatives.
This motivates adaptive sparsity rather than a fixed top-$k$ truncation.

To examine this hypothesis, we compare softmax and entmax under different
candidate-set sizes
$k\in\{2,4,8,16,32\}$.
For each $k$, we retain the top-$k$ logits and construct the soft token using
either softmax or entmax.
Softmax assigns non-zero probability to all retained candidates, whereas
entmax can further reduce the active support according to their relative
scores.
Therefore, $k$ controls the maximum candidate budget, while entmax determines
the effective degree of softness within this budget.
We train all variants on GSM8K using the same configuration as the main
experiments, except that the maximum response length is set to 2,048
tokens for computational efficiency, and record the mean Pass@1 over 4 rollouts at each validation.

Fig.~\ref{fig:topk-ablation} shows a clear difference between the two schemes.
With softmax, training becomes increasingly sensitive to $k$, and larger
candidate sets lead to substantial performance degradation.
In contrast, entmax improves the performance considerably as $k$ increases, showing the RL training benefits from a larger candidate pool without forcing all candidates to
participate in the soft token.
These results support our hypothesis that soft reasoning benefits from
adaptive rather than fixed sparsity: a larger candidate set provides
additional alternatives when needed, while entmax suppresses unnecessary
components at confident reasoning steps.

\section{Related Work}

\textbf{RL reasoning for LLMs.}
RL has been shown to be effective for improving LLM reasoning with Chain-of-Thought
\citep{shao2024deepseekmath,yu2025dapo,hu2025reinforcepp,zheng2025gspo}.
However, exploration through discrete token sampling is inefficient over the
large space of possible reasoning trajectories
\citep{liao2025exploration}.
This exploration bottleneck can be further exacerbated as RL training
suppresses valuable low-probability alternatives
\citep{huang2026lowprob}, which limits the diversity of reasoning trajectories explored by the model
\citep{wu2026ctrls,zhang2025softthinking}.

\textbf{Soft reasoning.}
Recent studies perform
reasoning in a continuous representation space so that a single
reasoning step can preserve information from multiple 
reasoning paths
\citep{hao2025coconut,xu2025softcot,wu2025pccot,shen2025codi}.
Theoretical analyses show that continuous reasoning representation can encode
multiple reasoning paths in superposition and track multiple search frontiers
in parallel, providing greater representational expressivity and more
efficient exploration than following a single discrete trajectory
\citep{zhu2025superposition,gozeten2026continuous}.

Building on this principle, continuous reasoning has been realized in various ways, such as
through internal hidden representations
\citep{hao2025coconut,xu2025softcot,wu2025pccot,shen2025codi} and through mixtures of token embeddings
\citep{zhang2025softthinking,wang2025softconcept,tang2026multiplex}.
We focus on the latter formulation, which constructs each soft token as a
mixture of vocabulary-token embeddings, preserving multiple token-level
alternatives while remaining anchored to the pretrained token embedding
space.
Subsequent work extends the simple weighted-average soft-token construction
through diversified test-time reasoning \citep{xu2025softcotpp},
uncertainty-aware activation \citep{fu2026selar, xu2026thinkrouter}, and stochastic
perturbations that mitigate soft-token collapse
\citep{wu2026singlethreaded}.
These works indicate the potential of soft-token-based
continuous reasoning for improving the expressiveness, efficiency, and exploration capability of LLM inference.

\textbf{RL training with soft CoT.}
While soft CoT improves reasoning at inference time, pretrained LLMs are
optimized over discrete token sequences and are never trained to
exploit continuous mixtures of token alternatives
\citep{rizvimartel2026illusion}, which limits the potential of soft reasoning.
This motivates post-training methods for soft reasoning.
Recent approaches introduce stochastic soft tokens through Gaussian
perturbations or Gumbel-based sampling and optimize them with RL objectives
\citep{butt2026softtokens,zheng2025softgrpo,zhou2026lepo}.
However, such optimization does not propagate reward credit through the differentiable dependence between consecutive
soft tokens.
SpaRLax fills this gap by reparameterizing soft-token generation
and propagating pathwise gradients through the soft-token sequence, allowing
an earlier soft-token distribution to receive credit according to how it
affects subsequent reasoning and the final answer.

\section{Conclusion}
We propose Sparse RL Relaxation for post-training soft reasoning with differentiable cross-step gradient propagation.
Our method further uses $\operatorname{entmax}$-based adaptive sparsity to retain useful alternatives while suppressing low-probability noise.
Experiments show consistent gains under both soft- and hard-CoT inference, indicating that improvements learned in the relaxed reasoning space can transfer effectively to discrete decoding.
Overall, Sparse RL Relaxation provides an effective framework for learning soft reasoning while remaining compatible with standard discrete inference.

\section*{Acknowledgments}
This research
was supported in part by the Natural Sciences and Engineering Research Council of Canada (NSERC), the Amii Fellow Program, the Canada CIFAR AI Chair Program, and the Digital
Research Alliance of Canada. This work was also supported in part by the project DECOLLAGE (ERC-2022-CoG 101088763) and by FCT/MECI through national funds and when applicable co-funded EU funds under UID/50008: Instituto de Telecomunicações.

\bibliographystyle{assets/plainnat}
\bibliography{references}

\beginappendix

\section{Implementation Details}
\label{app:implementation}

\subsection{Sparse RL Relaxation with GRPO}
\label{app:sparse_grpo}

We provide two implementations of Sparse RL Relaxation on the GRPO framework.
The first implements the full relaxation by differentiating through the
complete soft-CoT trajectory, while the second uses Proxy Gradient Grafting
for efficient training with decoupled rollout and optimization.
Both implementations use the Gumbel--entmax soft-token construction in
Eqn.~(\ref{eq:gumbel_entmax}) and differ only in how gradients through the
soft-token sequence are computed.

For a rollout group with rewards $\{R_g\}_{g=1}^{G}$, the group-relative
advantage is
\begin{align}
    \widehat A_g
    &=
    \frac{R_g-\mu_R}{\sigma_R+\varepsilon_A},
    \qquad
    \mu_R=\frac{1}{G}\sum_{h=1}^{G}R_h,
    \qquad
    \sigma_R=
    \sqrt{
        \frac{1}{G}
        \sum_{h=1}^{G}(R_h-\mu_R)^2
    }.
    \label{eq:impl_grpo_advantage}
\end{align}

Given token-level importance ratios $\{r_{g,j}\}$, define the shared clipped
GRPO loss
\begin{align}
    \mathcal L_{\mathrm{GRPO}}
    \left(
        \{r_{g,j}\},
        \{\widehat A_g\}
    \right)
    :=
    -
    \frac{1}{G}
    \sum_{g=1}^{G}
    \frac{1}{n_g}
    \sum_{j=1}^{n_g}
    \min
    \left\{
        r_{g,j}\widehat A_g,\,
        \operatorname{clip}
        \left(
            r_{g,j},
            1-\varepsilon_{\mathrm{clip}},
            1+\varepsilon_{\mathrm{clip}}
        \right)
        \widehat A_g
    \right\}.
    \label{eq:impl_grpo_loss}
\end{align}

We instantiate Sparse RL Relaxation on a common GRPO backbone.
The two implementations below differ only in the construction of the
current-policy soft-token gradient and the resulting importance ratio.

\paragraph{Sparse RL Relaxation with GRPO}
\label{app:full_sparse_grpo}

Algorithm~\ref{alg:full_sparse_grpo} differentiates through the complete
current-policy soft-CoT trajectory.

\begin{algorithm}[t]
\caption{Full Sparse RL Relaxation with GRPO}
\label{alg:full_sparse_grpo}
\begin{algorithmic}[1]
\Require
prompt batch $\mathcal B$,
policy $\pi_{\bm\theta}$,
group size $G$,
$\alpha$,
$\tau_g$,
learning rate $\eta$

\State $\bm\theta_{\mathrm{old}}\gets\operatorname{copy}(\bm\theta)$
\State $\mathcal L_{\mathrm{full}}\gets 0$

\For{$\bm x\in\mathcal B$}

    \For{$g=1,\ldots,G$}

        \For{$i=1,\ldots,m$}
            \State
            $\bm\epsilon_i^{(g)}
            \sim
            \mathrm{Gumbel}(0,1)^{|\mathcal V|}$

            \State
            $\bm z_i^{\mathrm{old},(g)}
            \gets
            f_{\bm\theta_{\mathrm{old}}}
            \left(
                \bm x,
                \tilde{\bm e}_{<i}^{\mathrm{old},(g)}
            \right)$

            \State
            $\tilde{\bm p}_i^{\mathrm{old},(g)}
            \gets
            \operatorname{entmax}_{\alpha}
            \left(
                \frac{
                    \bm z_i^{\mathrm{old},(g)}
                    +
                    \bm\epsilon_i^{(g)}
                }{\tau_g}
            \right)$

            \State
            $\tilde{\bm e}_i^{\mathrm{old},(g)}
            \gets
            \displaystyle
            \sum_{k=1}^{|\mathcal V|}
            \tilde p_{i,k}^{\mathrm{old},(g)}
            \bm e_k$
        \EndFor

        \State
        $\bm y_{1:n_g}^{(g)}
        \sim
        \pi_{\bm\theta_{\mathrm{old}}}
        \left(
            \cdot
            \mid
            \bm x,
            \tilde{\bm e}_{1:m}^{\mathrm{old},(g)}
        \right)$

        \State
        $R_g
        \gets
        R
        \left(
            \bm x,
            \bm y_{1:n_g}^{(g)}
        \right)$

        \State
        $\ell_{g,j}^{\mathrm{old}}
        \gets
        \log
        \pi_{\bm\theta_{\mathrm{old}}}
        \left(
            y_j^{(g)}
            \mid
            \bm x,
            \tilde{\bm e}_{1:m}^{\mathrm{old},(g)},
            \bm y_{<j}^{(g)}
        \right),
        \quad
        j=1,\ldots,n_g$

    \EndFor

    \State
    $\widehat A_g
    \gets
    \text{Eqn.~(\ref{eq:impl_grpo_advantage})},
    \quad
    g=1,\ldots,G$

    \For{$g=1,\ldots,G$}

        \For{$i=1,\ldots,m$}

            \State
            $\bm z_i^{\bm\theta,(g)}
            \gets
            f_{\bm\theta}
            \left(
                \bm x,
                \tilde{\bm e}_{<i}^{\bm\theta,(g)}
            \right)$

            \State
            $\tilde{\bm p}_i^{\bm\theta,(g)}
            \gets
            \operatorname{entmax}_{\alpha}
            \left(
                \frac{
                    \bm z_i^{\bm\theta,(g)}
                    +
                    \bm\epsilon_i^{(g)}
                }{\tau_g}
            \right)$

            \State
            $\tilde{\bm e}_i^{\bm\theta,(g)}
            \gets
            \displaystyle
            \sum_{k=1}^{|\mathcal V|}
            \tilde p_{i,k}^{\bm\theta,(g)}
            \bm e_k$

        \EndFor

        \State
        $\ell_{g,j}^{\bm\theta}
        \gets
        \log
        \pi_{\bm\theta}
        \left(
            y_j^{(g)}
            \mid
            \bm x,
            \tilde{\bm e}_{1:m}^{\bm\theta,(g)},
            \bm y_{<j}^{(g)}
        \right),
        \quad
        j=1,\ldots,n_g$

        \State
        $r_{g,j}^{\mathrm{full}}
        \gets
        \exp
        \left(
            \ell_{g,j}^{\bm\theta}
            -
            \ell_{g,j}^{\mathrm{old}}
        \right),
        \quad
        j=1,\ldots,n_g$

    \EndFor

    \State
    $\mathcal L_{\mathrm{full}}
    \gets
    \mathcal L_{\mathrm{full}}
    +
    \frac{1}{|\mathcal B|}
    \mathcal L_{\mathrm{GRPO}}
    \left(
        \{r_{g,j}^{\mathrm{full}}\},
        \{\widehat A_g\}
    \right)$

\EndFor

\State
$\bm\theta
\gets
\operatorname{OptimizerStep}
\left(
    \bm\theta,
    \nabla_{\bm\theta}\mathcal L_{\mathrm{full}};
    \eta
\right)$

\end{algorithmic}
\end{algorithm}

\paragraph{Efficient Sparse RL Relaxation with Proxy Gradient Grafting}
\label{app:proxy_sparse_grpo}

Algorithm~\ref{alg:proxy_sparse_grpo} replaces the full recursive gradient
with Proxy Gradient Grafting.

\begin{algorithm}[t]
\caption{Efficient Sparse RL Relaxation with Proxy Gradient Grafting}
\label{alg:proxy_sparse_grpo}
\begin{algorithmic}[1]
\Require
prompt batch $\mathcal B$,
policy $\pi_{\bm\theta}$,
group size $G$,
$\alpha$,
$\tau_g$,
learning rate $\eta$

\State $\bm\theta_{\mathrm{old}}\gets\operatorname{copy}(\bm\theta)$
\State $\mathcal L_{\mathrm{proxy}}\gets 0$

\For{$\bm x\in\mathcal B$}

    \For{$g=1,\ldots,G$}

        \For{$i=1,\ldots,m$}

            \State
            $\bm\epsilon_i^{(g)}
            \sim
            \mathrm{Gumbel}(0,1)^{|\mathcal V|}$

            \State
            $\bm z_i^{\mathrm{roll},(g)}
            \gets
            f_{\bm\theta_{\mathrm{old}}}
            \left(
                \bm x,
                \tilde{\bm e}_{<i}^{\mathrm{roll},(g)}
            \right)$

            \State
            $\tilde{\bm p}_i^{\mathrm{roll},(g)}
            \gets
            \operatorname{entmax}_{\alpha}
            \left(
                \frac{
                    \bm z_i^{\mathrm{roll},(g)}
                    +
                    \bm\epsilon_i^{(g)}
                }{\tau_g}
            \right)$

            \State
            $\tilde{\bm e}_i^{\mathrm{roll},(g)}
            \gets
            \displaystyle
            \sum_{k=1}^{|\mathcal V|}
            \tilde p_{i,k}^{\mathrm{roll},(g)}
            \bm e_k$

        \EndFor

        \State
        $\bm y_{1:n_g}^{(g)}
        \sim
        \pi_{\bm\theta_{\mathrm{old}}}
        \left(
            \cdot
            \mid
            \bm x,
            \tilde{\bm e}_{1:m}^{\mathrm{roll},(g)}
        \right)$

        \State
        $R_g
        \gets
        R
        \left(
            \bm x,
            \bm y_{1:n_g}^{(g)}
        \right)$

        \State
        $\ell_{g,j}^{\mathrm{old}}
        \gets
        \log
        \pi_{\bm\theta_{\mathrm{old}}}
        \left(
            y_j^{(g)}
            \mid
            \bm x,
            \tilde{\bm e}_{1:m}^{\mathrm{roll},(g)},
            \bm y_{<j}^{(g)}
        \right),
        \quad
        j=1,\ldots,n_g$

    \EndFor

    \State
    $\widehat A_g
    \gets
    \text{Eqn.~(\ref{eq:impl_grpo_advantage})},
    \quad
    g=1,\ldots,G$

    \For{$g=1,\ldots,G$}

        \For{$i=1,\ldots,m$}

            \State
            $\bm z_i^{\mathrm{proxy},(g)}
            \gets
            f_{\bm\theta}
            \left(
                \bm x,
                \operatorname{sg}
                \left(
                    \tilde{\bm e}_{<i}^{\mathrm{roll},(g)}
                \right)
            \right)$

            \State
            $\tilde{\bm p}_i^{\mathrm{proxy},(g)}
            \gets
            \operatorname{entmax}_{\alpha}
            \left(
                \frac{
                    \bm z_i^{\mathrm{proxy},(g)}
                    +
                    \bm\epsilon_i^{(g)}
                }{\tau_g}
            \right)$

            \State
            $\bar{\bm p}_i^{(g)}
            \gets
            \operatorname{sg}
            \left(
                \tilde{\bm p}_i^{\mathrm{roll},(g)}
            \right)
            +
            \tilde{\bm p}_i^{\mathrm{proxy},(g)}
            -
            \operatorname{sg}
            \left(
                \tilde{\bm p}_i^{\mathrm{proxy},(g)}
            \right)$

            \State
            $\bar{\bm e}_i^{(g)}
            \gets
            \displaystyle
            \sum_{k=1}^{|\mathcal V|}
            \bar p_{i,k}^{(g)}
            \bm e_k$

        \EndFor

        \State
        $\bar\ell_{g,j}^{\bm\theta}
        \gets
        \log
        \pi_{\bm\theta}
        \left(
            y_j^{(g)}
            \mid
            \bm x,
            \bar{\bm e}_{1:m}^{(g)},
            \bm y_{<j}^{(g)}
        \right),
        \quad
        j=1,\ldots,n_g$

        \State
        $r_{g,j}^{\mathrm{proxy}}
        \gets
        \exp
        \left(
            \bar\ell_{g,j}^{\bm\theta}
            -
            \ell_{g,j}^{\mathrm{old}}
        \right),
        \quad
        j=1,\ldots,n_g$

    \EndFor

    \State
    $\mathcal L_{\mathrm{proxy}}
    \gets
    \mathcal L_{\mathrm{proxy}}
    +
    \frac{1}{|\mathcal B|}
    \mathcal L_{\mathrm{GRPO}}
    \left(
        \{r_{g,j}^{\mathrm{proxy}}\},
        \{\widehat A_g\}
    \right)$

\EndFor

\State
$\bm\theta
\gets
\operatorname{OptimizerStep}
\left(
    \bm\theta,
    \nabla_{\bm\theta}\mathcal L_{\mathrm{proxy}};
    \eta
\right)$

\end{algorithmic}
\end{algorithm}

For clarity, Algorithms~\ref{alg:full_sparse_grpo} and
\ref{alg:proxy_sparse_grpo} present the full-vocabulary formulation and
omit optimizer-specific and shared GRPO regularization details. In practice, this may result in heavy memory burden. Thus, recent methods \citep{zheng2025softgrpo, zhou2026lepo, butt2026softtokens} typically select only top-$k$ candidate logits for soft token construction.
More details of hyperparameter of the implementation can be found in the next subsection.

\subsection{Implementation Configuration and Hyperparameters}
\label{app:imp_configs}

To ensure a controlled comparison, we use a unified training configuration
across all methods and keep all shared components fixed whenever applicable.
Our setup closely follows SofT-GRPO~\citep{zheng2025softgrpo}, including the
GRPO backbone, rollout configuration, optimization settings, and training
budget. Competing methods differ only in hyperparameters required by their
respective soft-CoT mechanisms. Full configurations are reported in
Table~\ref{tab:hyperparameters}.

For SpaRLax, we set the entmax parameter to $\alpha=1.5$.
This is a commonly used choice in prior entmax-related papers
\citep{peters2019sparse,correia2019adaptively,tezekbayev2022speeding},
providing an intermediate level of sparsity between softmax
($\alpha=1$) and sparsemax ($\alpha=2$).
Importantly, $\alpha=1.5$ also admits a specialized exact algorithm,
making it particularly convenient to implement and with lower latency compared
with general values of $\alpha$, which typically require iterative
computation \citep{peters2019sparse, tezekbayev2022speeding}.

In addition, we use a candidate-set size of $k=32$ for SpaRLax for lower memory usage; future work can consider implementations with fused kernels~\citep{goncalves2025adasplash}.
Unlike softmax, entmax can adaptively reduce its effective support by
assigning exactly zero probability to unnecessary candidates.
This allows us to expose the model to a larger candidate pool without
forcing all retained candidates to participate in the soft-token mixture.
As shown in \S\ref{sec:sparsity_analysis}, softmax becomes increasingly
sensitive to large $k$ (which may be the reason why other methods \citep{zheng2025softgrpo, butt2026softtokens} select a modest $k$ in their work), whereas entmax remains substantially more robust
and can benefit from a larger candidate budget.
We therefore use $k=32$ for SpaRLax while retaining the smaller candidate
sets used by the softmax-based baselines.

Apart from these method-specific choices, all
shared training configurations are held fixed across methods.

\begin{table*}[t]
\centering
\resizebox{\textwidth}{!}{%
\begin{tabular}{lccccc}
\toprule
Parameter
& \shortstack{Discrete GRPO\\\citep{guo2025deepseekr1}}
& \shortstack{SofT-GRPO\\\citep{zheng2025softgrpo}}
& \shortstack{SpaRLax\\\ (ours)}
& \shortstack{STHT\\\citep{butt2026softtokens}}
& \shortstack{LEPO\\\citep{zhou2026lepo}} \\
\midrule

\multicolumn{6}{l}{\textit{Batch configuration}} \\
Fine-tuning & \multicolumn{5}{c}{Full-parameter fine-tuning} \\
Filter overlong prompts & Enabled & Enabled & Enabled & Enabled & Enabled \\
Prompt batch size & 32 & 32 & 32 & 32 & 16 \\
\shortstack[l]{Responses per prompt\\(group size)}
& 8 & 8 & 8 & 8 & 8 \\
\shortstack[l]{Global rollout batch\\(trajectories)}
& 256 & 256 & 256 & 256 & 128 \\
\shortstack[l]{Effective PPO minibatch\\(trajectories)}
& 256 & 256 & 256 & 256 & 128 \\
\shortstack[l]{Microbatch per GPU\\(trajectories)}
& 1 & 1 & 1 & 1 & 1 \\
Gradient accumulation steps & 64 & 64 & 64 & 64 & 64 \\

\midrule
\multicolumn{6}{l}{\textit{Optimization and regularization}} \\
Optimizer & AdamW & AdamW & AdamW & AdamW & AdamW \\
Learning rate & $10^{-6}$ & $10^{-6}$ & $10^{-6}$ & $10^{-6}$ & $10^{-6}$ \\
Adam $(\beta_1,\beta_2)$ & \multicolumn{5}{c}{$(0.9,0.999)$} \\
Weight decay & 0.01 & 0.01 & 0.01 & 0.01 & 0.01 \\
LR schedule / warmup steps & \multicolumn{5}{c}{Constant / 0} \\
PPO epochs per rollout & 1 & 1 & 1 & 1 & 2 \\
PPO clipping interval & \multicolumn{5}{c}{$[0.8,1.2]$} \\
Dual-clip coefficient & 3.0 & 3.0 & 3.0 & 3.0 & 3.0 \\
Gradient-norm clipping & 1.0 & 1.0 & 1.0 & 1.0 & 0.5 \\
\shortstack[l]{Group-standard-deviation\\normalization}
& Enabled & Enabled & Enabled & Enabled & Enabled \\
Entropy bonus coefficient & 0 & 0 & 0 & 0 & 0 \\
Actor KL coefficient & $10^{-3}$ & $10^{-3}$ & $10^{-3}$ & $10^{-3}$ & $10^{-3}$ \\
LEPO KL estimator & -- & -- & -- & -- & $1$ \\
\shortstack[l]{Max training steps\\(outer iterations)}
& $3,000$ & $3,000$ & $3,000$ & $3,000$ & $3,000$ \\

\midrule
\multicolumn{6}{l}{\textit{Training-time generation and method configuration}} \\
Soft thinking & Disabled & Enabled & Enabled & Enabled & Enabled \\
Soft-token transform & -- & Softmax & Entmax & Softmax & Softmax \\
Soft candidate count & -- & 5 & 32 & 5 & 30 \\
Entmax $\alpha$ & -- & -- & 1.5 & -- & -- \\
Relaxation temperature & -- & 0.1 & 1.0 & 1.0 & 0.5 \\
Gumbel-noise multiplier & -- & 1.0 & 1.0 & 1.0 & 1.0 \\
\shortstack[l]{Categorical sampling\\temperature}
& 1.0 & 1.0 & 1.0 & 1.0 & 1.0 \\
\shortstack[l]{Categorical top-$k$\\ / top-$p$}
& 5 / 0.95 & 5 / 0.95 & 5 / 0.95 & 5 / 0.95 & 5 / 0.95 \\
Direct PG positions & All & All & answer phase & All & answer phase \\
LEPO maximum latent steps & -- & -- & -- & -- & 64 \\
STHT density scale $\sigma$ & -- & -- & -- & $0.33\,\mathrm{RMS}(E)$ & -- \\
STHT log-score range & -- & -- & -- & $[-50,0]$ & -- \\

\midrule
\multicolumn{6}{l}{\textit{Precision and parallelism}} \\
Forward precision & BF16 & BF16 & BF16 & BF16 & BF16 \\
GPUs & \multicolumn{5}{c}{4 NVIDIA H100 GPUs} \\
Actor sharding & FSDP1 & FSDP1 & FSDP1 & FSDP1 & FSDP1 \\
\shortstack[l]{Sequence / rollout tensor\\parallel size}
& 1 / 1 & 1 / 1 & 1 / 1 & 1 / 1 & 1 / 1 \\
Gradient checkpointing & Enabled & Enabled & Enabled & Enabled & Enabled \\
\shortstack[l]{Parameter / optimizer\\offload}
& Enabled & Enabled & Enabled & Enabled & Enabled \\

\bottomrule
\end{tabular}%
}
\caption{Training hyperparameters. The first four columns retain the supplied GSM8K settings; LEPO reports launcher defaults, which use DeepScaleR training data. Batch sizes distinguish prompt groups from generated trajectories.}
\label{tab:hyperparameters}
\end{table*}

\subsection{Stopping Criterion for Soft Reasoning}
\label{app:stopping_criterion}

We follow the stopping criterion used in prior soft-thinking methods
\citep{zhang2025softthinking,zheng2025softgrpo} to determine when to
transition from soft reasoning back to standard discrete decoding.
At each reasoning step $i$, after constructing the soft-token distribution
$\tilde{\bm p}_i$, we identify its most probable vocabulary token as
\begin{align}
    t_i^\star
    =
    \arg\max_{k\in\mathcal V}
    \tilde p_{i,k}.
\end{align}
Soft reasoning continues while
$t_i^\star \neq \langle/\mathrm{think}\rangle$.
Once the end-of-thinking token becomes the most probable token, i.e.,
$t_i^\star=\langle/\mathrm{think}\rangle$, we terminate the soft reasoning
phase and switch to standard autoregressive decoding for the final answer.
All subsequent answer tokens are discrete and sampled from the
model's vocabulary in the usual manner.

This stopping rule is applied during both rollout and inference and does
not introduce an additional learned stopping module or a method-specific
stopping criterion.

\section{Limitations and Future Work}
\label{app:limitations}

\paragraph{Superposition utilization.}
Our analysis in \S\ref{sec:superposition_analysis} provides evidence that
the non-top-1 components of a soft token have a measurable effect on
downstream model computation, as our SpaRLax exhibits a stronger effect than the
competing methods.
This intervention provides an indirect measure of superposition
utilization, as it does not fully show that multiple trajectories are truly used during exploration. In future work, we are interested in exploring how such superposition can help search tasks, such as the Traveling Salesperson Problem.

\paragraph{Scale and generalization.}
Although we evaluate the trained models across nine benchmarks spanning
mathematical, knowledge-intensive, and code reasoning, our post-training
experiments are conducted with one base model and one RL training
corpus. However, we cannot afford more experiments using assorted LLMs and corpora. Nevertheless, we have carefully designed the experiments by isolating the effect of the proposed learning mechanism from the underlying model, training
data, and RL algorithm, as the configurations are fixed across all competing
methods. Moreover, the consistent improvements observed across multiple
reasoning domains suggest that the benefit is not confined to a specific setup.
We leave the scaling of this study to large technology companies.
\end{document}